\documentclass[10pt,twocolumn,letterpaper]{article}

\usepackage{iccv}
\usepackage{times}
\usepackage{epsfig}
\usepackage{graphicx}
\usepackage{amsmath}
\usepackage{amssymb}
\usepackage{multirow}
\usepackage[utf8]{inputenc}
\usepackage{pifont}
\usepackage{array}
\usepackage[hang,flushmargin]{footmisc} 
\usepackage[pagebackref=true,breaklinks=true,letterpaper=true,colorlinks,bookmarks=false]{hyperref}

\iccvfinalcopy 

\def\iccvPaperID{9} 

\ificcvfinal\fi
\begin{document}

\title{Sequentially Aggregated Convolutional Networks}

\author{Yiwen Huang$^*$$^{\dag}$$^{1}$ \ \  Pinglai Ou$^*$$^{2}$ \ \  Rihui Wu$^*$$^{3}$ \ \   Ziyong Feng$^{4}$ \\
        $^{1}$Wenhua College, Huazhong University of Science and Technology \ \  $^{2}$Virginia Tech \\
        $^{3}$University of Sydney \ \  $^{4}$DeepGlint Technology Limited \\
    {\tt\small nickgray0@gmail.com, kice@vt.edu, wrhkurt@gmail.com, ziyongfeng@deepglint.com}
}

\maketitle

\ificcvfinal\thispagestyle{empty}\fi

\begin{abstract}
    Modern deep networks generally implement a certain form of shortcut connections to alleviate optimization difficulties. However, we observe that such network topology alters the nature of deep networks. In many ways, these networks behave similarly to aggregated wide networks. We thus exploit the aggregation nature of shortcut connections at a finer architectural level and place them within wide convolutional layers. We end up with a sequentially aggregated convolutional (SeqConv) layer that combines the benefits of both wide and deep representations by aggregating features of various depths in sequence. The proposed SeqConv serves as a drop-in replacement of regular wide convolutional layers and thus could be handily integrated into any backbone network. We apply SeqConv to widely adopted backbones including ResNet and ResNeXt, and conduct experiments for image classification on public benchmark datasets. Our ResNet based network with a model size of ResNet-50 easily surpasses the performance of the 2.35$\times$ larger ResNet-152, while our ResNeXt based model sets a new state-of-the-art accuracy on ImageNet classification for networks with similar model complexity. The code and pre-trained models of our work are publicly available at \url{https://github.com/GroupOfAlchemists/SeqConv}.
\end{abstract}

\vspace{-9mm}
\let\thefootnote\relax\footnote{$^*$ Equal contribution.}
\let\thefootnote\relax\footnote{$^{\dag}$ This work was done when Yiwen Huang was an intern at DeepGlint Technology Limited.}
\section{Introduction}
\label{sec:introduction}

\begin{figure*}[t]
    \begin{center}
        \includegraphics[width=1.0\textwidth]{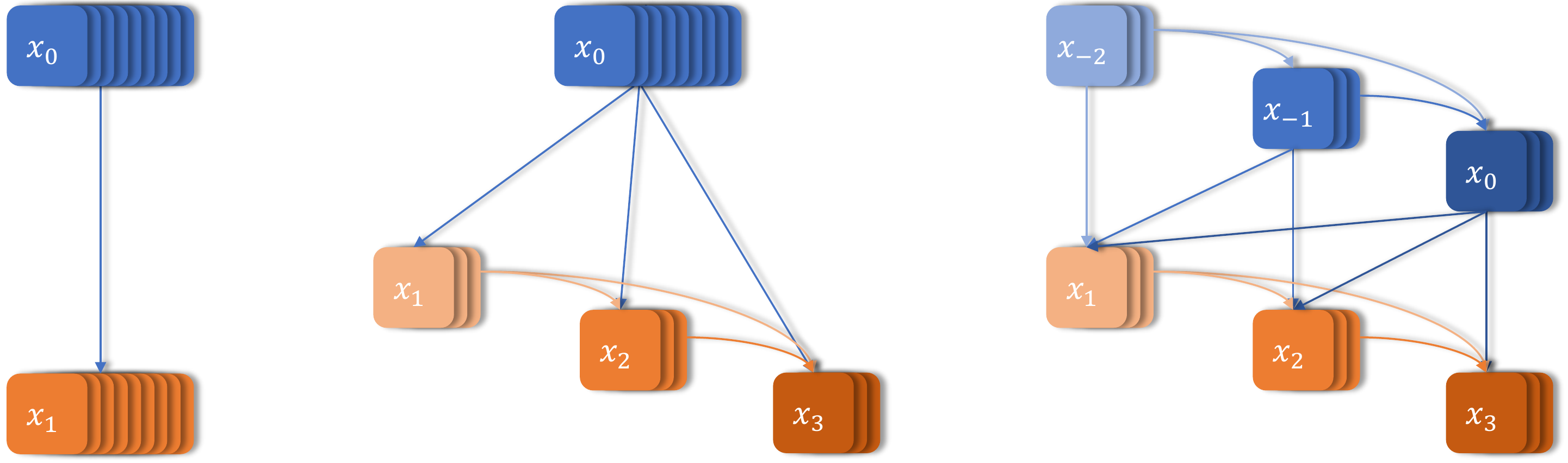}
        \caption{A regular wide convolutional layer (left), a sequentially aggregated convolutional layer (middle) and its windowed variant (right) with $g$=$3$.}
        \label{fig.1}
        \vspace{-3 ex}
    \end{center}
\end{figure*}

Convolutional neural networks (CNNs) have gained overwhelming success for visual recognition owing to their representational ability. Recent work has shown that the depth of representation is of crucial importance to the performance of CNNs~\cite{simonyan2014very, szegedy2015going, he2016deep, huang2017densely}. Eldan~\etal conclude that depth is a determinant factor of the expressiveness of neural networks~\cite{eldan2016power}. Several studies~\cite{lu2017expressive, zagoruyko2016wide, arora2019exact} have also been conducted to investigate the width of the representation, it however does not seem to be the major concern of recent network architecture designs. The possibility to utilize the representational power of both wide and deep representations under a given model complexity has remained an unexplored problem.

Increasing depth by simply stacking more layers leads to optimization difficulties as the information flow gets gradually obscured by each layer during propagation~\cite{bengio1994learning, srivastava2015highway} in deep networks. An intuitive approach to ameliorate this problem is introducing shortcut connections towards farther layers to enable direct access to the guiding signal through propagation. This method has been shown particularly effective by various recently proposed state-of-the-art networks~\cite{he2016deep, larsson2017fractalnet, huang2017densely, chen2017dual}, with its effectiveness further confirmed by visualizing the loss landscape of such networks~\cite{li2018visualizing}.

Despite the fact that shortcut connections make it viable to optimize extremely deep networks as they help preserve the information flow, they seem to change the expected behaviors of deep networks. In many ways, these networks exhibit the property of having weak dependencies between consecutive layers while layers generally share strong dependencies in regular deep networks~\cite{veit2016residual}. It is reported that ResNets~\cite{he2016deep}, a typical network architecture with heavy use of shortcut connections, behave similarly to ensembles of many shallow networks~\cite{veit2016residual}, suggesting that such networks could be viewed from the aspect of a collection of several mostly independent subsections. This quality of having independent subsections and the aggregation nature of skip connections provide us the insight to link networks with shortcut connections to pseudo-wide networks.

The benefit of having wide representations lies in the fact that it allows for a larger feature space by introducing higher feature throughput to the network, however we argue that wide convolutional layers are not the only means to achieve this goal. A wide representation could also be collectively formulated by aggregating many transformations with small kernels. The shortcut connections are a case of aggregated transformations as they aggregate outputs from many layers, thus there is no surprise when we observe certain properties that resemble the behavior of a wide network on a deep and thin network with shortcut connections.

We propose a novel aggregation-based convolutional layer (SeqConv) to construct networks with the benefits of both wide and deep representations following the aggregation nature of shortcut connections. We divide a regular wide convolutional layer into several groups and place in-layer shortcut connections between each group. We then aggregate the outputs from all groups in sequence to formulate a collective wide representation. The SeqConv layer not only preserves the width of a regular convolutional layer, but as well introduces a hierarchical multi-path micro-architecture that is capable of representing heterogeneous kernels~\cite{singh2019hetconv}. The representation capability of the layer is thus greatly enhanced, it is possible for a single SeqConv layer to produce multi-scale representation~\cite{szegedy2015going, szegedy2017inception} with deep hierarchical features. Our contributions in this paper are threefold:

\begin{itemize}
    \item We propose sequentially aggregated convolutional (SeqConv) layers, along with several enhanced variants, that are capable of producing stronger representations than standard convolutional layers.

    \item We analyze the relations of SeqConv to DenseNet~\cite{huang2017densely}, and reinterpret the success of DenseNet with small growth rate from the perspective of sequentially aggregated wide representations. A windowed aggregation mechanism is also proposed to address the parameter redundancy and high computational cost of dense aggregation.

    \item  We adopt SeqConv as the drop-in replacement of regular wide convolutional layers to construct networks for image classification. Our models achieve higher accuracy than significantly larger state-of-the-art models.
\end{itemize}

\section{Related Work}
\label{sec:related_work}

\vspace{-3 ex}

\noindent \paragraph{Skip connectivity.} 
Deep networks have been shown hard to optimize via gradient-based methods due to obstructed information flow through propagation, namely the diminishing feature reuse problem for forward propagation~\cite{srivastava2015highway} and the vanishing gradient problem for backward propagation~\cite{bengio1994learning}. Networks with shortcut connections~\cite{he2016deep, larsson2017fractalnet, huang2017densely, chen2017dual} were proposed to alleviate such optimization difficulties by introducing shorter paths to farther layers and thus preserving the information flow through propagation. Several implementations of skip connectivity have been proposed to demonstrate the effectiveness of this network topology. Highway networks~\cite{srivastava2015highway, srivastava2015training} and residual networks~\cite{he2016deep, he2016identity} construct skip connections with addition. Fractal networks~\cite{larsson2017fractalnet} replace addition with element-wise mean which makes no distinction between signals from each path and thus allows a new form of regularization, Drop-path, to be applied. DenseNets~\cite{huang2017densely} and its successors~\cite{huang2018condensenet, zhu2018sparsely} adopt concatenation to implement shortcut connections and attain favorable performance over previous work.

\vspace{-2 ex} \paragraph{Ensembles of relatively shallow networks.}
Several analyses were conducted to investigate properties of the particularly effective residual networks. Veit~\etal~\cite{veit2016residual} reported that removing building blocks from residual networks or only keeping the shortcut paths did not lead to apparent accuracy drop. Huang~\etal~\cite{huang2016deep} randomly dropped residual blocks while training and actually obtained improved performance. Both studies suggest that layers in residual networks do not share strong dependencies between each other and such observation is not expected for a regular deep network. As reported by Veit~\etal~\cite{veit2016residual}, removing layers from a VGG network does lead to drastic performance drop. This indicates that a residual network does not actually exhibit behaviors of an ultra-deep network, it rather behaves similarly to ensembles of many mutually independent shallow networks.

\vspace{-2 ex} \paragraph{Width \textit{vs}. Depth for ResNets.}
As discussed in Section~\ref{sec:introduction} that the aggregation nature of shortcut connections links deep networks with such topology to pseudo-wide networks, we compare residual aggregation with actual wide networks. We find that simply widening the network is of higher efficiency than stacking more residual blocks once the network has reached a certain depth. Reported by Zagoruyko~\etal~\cite{zagoruyko2016wide}, a 40-layer residual network of 4$\times$ width outperformed a 1001-layer network on CIFAR with fewer parameters. A wider 101-layer residual network also achieved higher accuracy on ImageNet classification than a 200-layer network with the same model complexity~\cite{xie2017aggregated}. One possible explanation is that residual aggregation entangles outputs from each layer and thus hinders the ability to search for new features~\cite{zhu2018sparsely}. We hence implement in-layer shortcut connections for SeqConv with concatenation instead of addition to avoid such limitation.

\vspace{-2 ex} \paragraph{Aggregated transformations.}
The implementation of aggregated transformations is generally supported by a multi-path architecture. Each path applies a transformation with a small kernel and features produced by each path are then aggregated to formulate the final representation in a larger feature space. The representation capability is determined by the multi-path architecture and this could be, a set of homogeneous paths~\cite{xie2017aggregated}, a set of hierarchical paths~\cite{huang2017densely}, other more complex structures~\cite{yu2018deep, szegedy2017inception}, or even learnable structures as reflected by the cell design of recent studies on network architecture search~\cite{zoph2018learning, huang2018gpipe}.

\section{Methods}

\subsection{Sequentially Aggregated Transformations}

Consider a regular wide convolutional layer gets divided into several groups of transformations, we employ a simple yet elegant hierarchical multi-path architecture to aggregate each group as briefly described in Section~\ref{sec:introduction}. A comparison between a standard convolutional layer and the proposed sequentially aggregated convolutional (SeqConv) layers is illustrated in Figure~\ref{fig.1}.

\vspace{-2 ex} \paragraph{Basic layers.}
For a SeqConv layer with $g$ groups of transformations, let $x_0$ denote the input of the layer, $x_i$ denote the output of the $i^{th}$ group. The layer is defined by:

\begin{equation}
    \label{eq:1}
    x_i = F_i([x_0, x_1, \dotsc, x_{i-1}])
\end{equation}

\noindent where $F_i$ denotes the non-linear transformation function of the $i^{th}$ group, while $[\dotsc]$ refers to the concatenation operation. The final representation produced by a SeqConv layer is formulated by the aggregation of outputs from all $g$ groups $[x_1, x_2, \dotsc,x_g]$. The width of the representation is preserved by the concatenation-based aggregation while the depth is drastically increased. For each transformation function, the sequential aggregation enables the view over features extracted by previous groups. Each time a group of features $x_i$ pass through the transformation function $F_{i+1}$ of the following group, a group of deeper features $x_{i+1}$ could be extracted. The final representation aggregates features of various depths, including very deep features from latter groups of the layer. The representation capability is thus greatly enhanced~\cite{eldan2016power}.

\begin{figure*}[t]
    \begin{center}
        \includegraphics[width=1.0\textwidth]{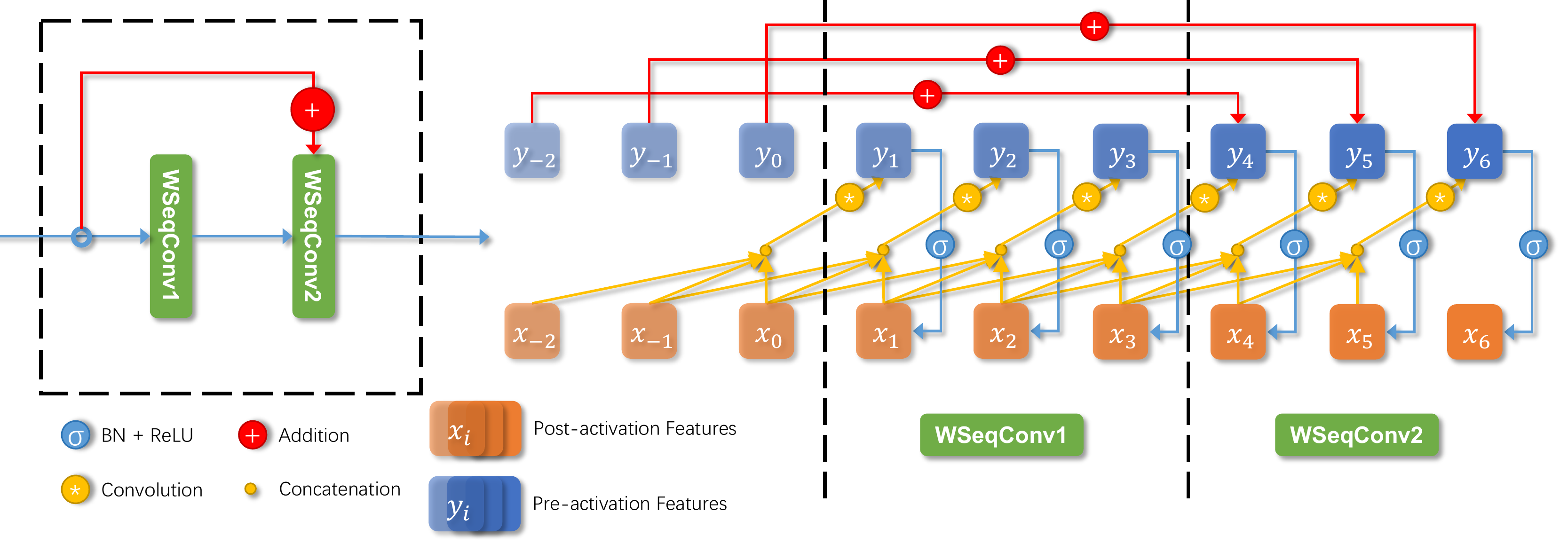}
        \caption{A residual block (left) and an exploded view of the block (right) revealing its internal structure in details.}
        \label{fig.2}
        \vspace{-3 ex}
    \end{center}
\end{figure*}

\vspace{-2 ex} \paragraph{Relations to DenseNet.}
The sequential aggregation was first introduced to convolutional neural networks by DenseNet~\cite{huang2017densely}. The hierarchical multi-path architecture defined by Eq.\ref{eq:1} is shared by both SeqConv and a dense block in DenseNet, except that SeqConv does not include the input $x_0$ in its final representation similar to a regular convolutional layer while a dense block does. Aside from the apparent architectural similarity, SeqConv is still fundamentally different from DenseNet in two aspects:
\begin{itemize}

\item SeqConv is derived on a different basis than that of DenseNet. In~\cite{huang2017densely}, Huang~\etal place heavy emphasis on feature reuse and improved information flow for deep networks using shortcut connections. SeqConv, instead, is based on the observation that it is not viable to build genuine deep networks using such network topology due to the vanishing gradient problem on the longest gradient path~\cite{veit2016residual}. The aggregation nature and shorter gradient paths of shortcut connections already lead to behaviors resembling wide architectures in a seemingly deep network. We thus embrace this side effect that links shortcut connections to wide architectures to modify \emph{actual} wide networks. Concretely, we utilize the hierarchical aggregation capability of shortcut connections as formulated by SeqConv to enhance the representational power of a single wide convolutional layer.

\item SeqConv is derived from wide convolutional layers, and thus has a finer architectural granularity than DenseNet. SeqConv is a \textbf{layer-level} architecture and could be integrated into a large variety of backbone networks such as ResNet~\cite{he2016identity}, ResNeXt~\cite{xie2017aggregated}, DLA~\cite{yu2018deep},~\etc by simply replacing the regular convolutional layers. Such flexibility has not been explored in~\cite{huang2017densely} since each transformation unit in DenseNet is regarded as a separate layer. We argue that such interpretation not only impedes the flexibility to integrate sequential aggregation into other backbone networks, but itself might also be problematic. We further analyze this limitation in our following reinterpretation of DenseNet.
\end{itemize}

The layers in DenseNet are unusually narrow, the only rationale seems to be a vague statement about “collective knowledge”~\cite{huang2017densely} and a clear analysis is absent. A layer, as described by~\cite{lecun2015deep}, produces a representation of the input. This clearly is not the case of a “layer” in DenseNet. Features extracted by a DenseNet “layer” are not enough to solely constitute a representation and are instead always amalgamated with features extracted by other “layers” to jointly comprise a hierarchical representation. It thus might not be accurate to refer to such transformation units in DenseNet as “layers”, since each one of them only contributes to one feature group of a very wide representation that comprises many such groups. We attribute the success of sequential aggregation, including DenseNet, to the expressiveness of hierarchical wide representations. Reuse of feature groups~\cite{huang2017densely} is an implementation to produce such representations. Moreover, overemphasis on feature reuse also has its own drawbacks as we address in the following section of windowed SeqConv layers. It is also worth noting that DenseNet bears a close resemblance to SeqConv plugged into a VGG-like~\cite{simonyan2014very} backbone network.

\vspace{-2 ex} \paragraph{Transformation functions.} 
For a basic SeqConv layer, we follow the common settings in~\cite{he2016deep} and implement the transformation function with three consecutive operations: 3$\times$3 convolution (Conv) followed by batch normalization (BN)~\cite{ioffe2015batch} and a rectified linear unit (ReLU)~\cite{glorot2011deep}. The number of filters for each group is determined by the channel number of the convolution and is denoted by $k$.

To further improve the computational efficiency and model compactness, we also implement a bottleneck transformation function following~\cite{he2016deep, szegedy2016rethinking, huang2017densely}. For this bottleneck variant, we first employ a 1$\times$1 Conv to reduce the channel number of the aggregated features down to $k$, then apply the transformation with a 3$\times$3 Conv. The bottleneck transformation function comprises six consecutive operations: Conv(1$\times$1)-BN-ReLU-Conv(3$\times$3)-BN-ReLU, and the SeqConv layer with this transformation function is marked by the “B” postfix (SeqConvB).

\vspace{-2 ex} \paragraph{Windowed layers.} 
The dense aggregation defined by Eq.\ref{eq:1} assigns more weights to earlier features of a representation produced by SeqConv. Consider a representation $x'_0$ of $g'$ groups $[x_{1-g'}, x_{2-g'}, \dotsc , x_0]$ produced by a previous SeqConv layer goes through the current layer. The earliest features $x_{1-g'}$ are utilized by both $F_{2-g'}, \dotsc, F_0$ of the previous layer and all transformation functions of the current layer, while the latest features $x_0$ are only utilized by the current layer. $x_{1-g'}$ is thus assigned with more weights than $x_0$ since each time a group of features pass through a transformation function, certain weights are assigned to that group of features. The extra weights assigned to earlier features give rise to a vast number of required parameters growing at an asymptotic rate of $O(n^2)$, where $n$ is the width of the SeqConv layer, whereas a regular convolutional layer merely has a linear parameter growth rate. Recent study~\cite{zhu2018sparsely} suggests that this quadratic growth suffers from significant parameter redundancy. It is observed that DenseNet, which shares the same aggregation mechanism with SeqConv, has many skip connections with average absolute weights close to zero~\cite{zhu2018sparsely, huang2017denselyc}. We also notice that features exploited by a particular group are mostly distributed over the outputs of recent preceding groups of that group~\cite{huang2018condensenet, zhu2018sparsely}, since the information carried by the outputs of earlier groups has been abundantly exploited.

Thus, to reduce the parameter redundancy and lower the computational cost of SeqConv, we propose a windowed variant of SeqConv (WSeqConv) that only aggregates the outputs from most recent groups. The WSeqConv is defined as follows:

\begin{equation}
    \label{eq:2}
    x_i = F_i([x_{i-g'}, x_{i-g'+1}, \dotsc, x_{i-1}])
\end{equation}

\noindent the representation produced by WSeqConv is still formulated by the aggregation $[x_1, x_2, \dotsc, x_g]$. Note that Eq.\ref{eq:2} is equivalent to applying a sliding rectangular window function $\phi$ across the channel dimension on Eq.\ref{eq:1}:

\begin{equation}
    \label{eq:3}
\phi_i(x) = 
    \begin{cases}
        1 & i - g' \leq x \leq i - 1 \\
        0 & \text{otherwise}
    \end{cases}
\end{equation}

\begin{equation}
    \label{eq:4}
    \omega_i = [\phi_i(1 - g'), \phi_i(2 - g'), \dotsc, \phi_i(i - 1)]
\end{equation}

\begin{equation}
    \label{eq:5}
    \begin{aligned}
        x_i &= F_i([x_{i-g'}, x_{i-g'+1}, \dotsc, x_{i-1}]) \\
            &= F_i([x_{1-g'}, x_{2-g'}, \dotsc, x_{i-1}] \circ \omega_i) \\
            &= F_i([x'_0, x_1, \dotsc, x_{i-1}] \circ \omega_i)
    \end{aligned}
\end{equation}

\noindent $\circ$ denotes the operator for the Hadamard product. The window $\phi$ truncates the input for each group to a constant width, the parameter number and computational cost of WSeqConv are thus reduced to the same as of a regular convolutional layer.

\subsection{Network Architecture}
\label{sec:network_architecture}
We apply SeqConv and WSeqConv to three widely-adopted backbone networks, pre-activation ResNet~\cite{he2016identity} with basic residual blocks, ResNet with bottleneck residual blocks and ResNeXt~\cite{xie2017aggregated} and refer to them as, respectively, SeqResNet, SeqResNet-B and SeqResNeXt. We evaluate these models on image classification following~\cite{he2016deep, xie2017aggregated, huang2017densely, zhu2018sparsely}.

\vspace{-2 ex} \paragraph{Residual blocks.}
We construct residual blocks with WSeqConv layers (or its bottleneck variant) as the building blocks of our networks. Following~\cite{he2016deep}, we place two layers in each residual block. Residual connections are embedded in the aggregation of the second layer owing to pre-activation identity mapping as in~\cite{he2016identity}. This specific residual connectivity pattern allows earlier features that fall out of the aggregation view of the second layer to be implicitly shared with the layer without introducing any extra parameters, which further encourages feature reuse at the network level. A detailed breakdown of the structure of residual blocks with WSeqConv layers is illustrated in Figure~\ref{fig.2}.

\begin{figure}[t]
    \begin{center}
        \includegraphics[width=0.5\textwidth]{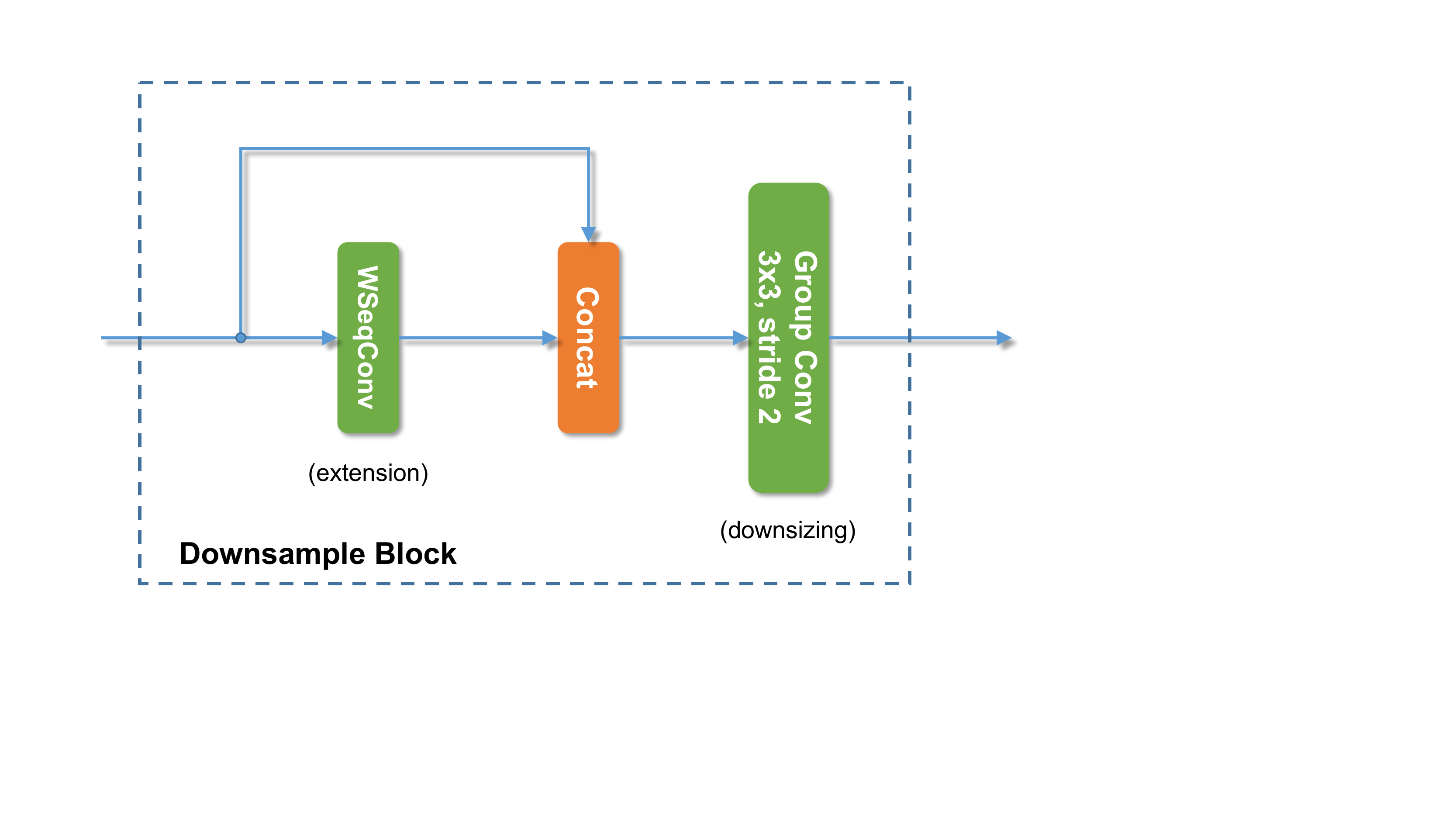}
        \caption{The topology of a down-sampling block, the extension layer and downsizing layer are, respectively, implemented by a WSeqConv layer and a grouped Conv layer of stride 2.}
        \label{fig.3}
        \vspace{-1.5 ex}
    \end{center}
\end{figure}

\vspace{-2 ex} \paragraph{Down-sampling blocks.}
Down-sampling is an essential part of the classification networks as it enables the network to extract features from different levels of abstraction. The common practice of down-sampling in a classification network reduces the spatial resolution of each feature map while the width of the representation is multiplied. This is however incompatible with sequential aggregation since it is not viable to aggregate feature maps of different spatial dimensions. We thus introduce down-sampling blocks to facilitate down-sampling for networks with SeqConv. A down-sampling block consists of an extension layer and a downsizing layer as illustrated in Figure~\ref{fig.3}. The representation is first extended to the target width by the extension layer. The spatial dimensions are then reduced by the downsizing layer. The grouping of downsizing prevents information leaks among features of different groups and hence preserves the hierarchy of the aggregated features, which is essential if the down-sampling block is followed by a WSeqConv layer that requires a hierarchical input.

\vspace{-2 ex} \paragraph{Subgroups.}
A new dimension called “cardinality” was introduced in ResNeXt\cite{xie2017aggregated}. This dimension divides a 3$\times$3 Conv into many small groups and allows for a wider representation compared to a regular convolutional layer with the same model complexity. To facilitate cardinality for SeqConv, we further divide the 3$\times$3 Conv in the transformation function of each group into several subgroups by replacing it with a 3$\times$3 grouped Conv. The number of subgroups for each group is denoted by $c$.

\newcommand{\blockseqc}[1]{\multirow{3}{*}{
\(\left[
\begin{array}{l}
\text{3$\times$3, #1 $\times$ $r$ \textcolor{blue}{*}}\\
[-.1em] \text{3$\times$3, #1 $\times$ $r$ \textcolor{blue}{*}}
\end{array}\right]\)$\times$ $N$}
}

\newcommand{\blockseqcb}[1]{\multirow{3}{*}{
\(\left[
\begin{array}{l}
\text{1$\times$1$\to$3$\times$3, #1 $\times$ $r$ \textcolor{blue}{*}}\\
[-.1em] \text{1$\times$1$\to$3$\times$3, #1 $\times$ $r$ \textcolor{blue}{*}}
\end{array}\right]\)$\times$ $N$}
}

\newcommand{\blockb}[3]{\multirow{3}{*}{
\(\left[
\begin{array}{l}
\text{1$\times$1, #2}\\
[-.1em] \text{3$\times$3, #2}\\
[-.1em] \text{1$\times$1, #1}
\end{array}\right]\)$\times$#3}
}

\newcommand{\blockx}[3]{\multirow{3}{*}{
\(\left[
\begin{array}{l}
\text{1$\times$1, #2}\\
[-.1em] \text{3$\times$3, #2, 32 groups}\\
[-.1em] \text{1$\times$1, #1}\\
\end{array}\right]\)$\times$#3}
}

\newcommand{\blockdense}[3]{\multirow{3}{*}{
\(\left[
\begin{array}{l}
\text{1$\times$1, #2}\\
[-.1em] \text{3$\times$3, #1}
\end{array}\right]\)$\times$#3}
}

\newcommand{\blockseq}[3]{\multirow{3}{*}{
\(\left[
\begin{array}{l}
\text{1$\times$1$\to$3$\times$3, #1, $k$=#2 \textcolor{blue}{*}}\\
[-.1em] \text{1$\times$1$\to$3$\times$3, #1, $k$=#2 \textcolor{blue}{*}}
\end{array}\right]\)$\times$#3}
}

\newcommand{\ft}[1]{\fontsize{#1pt}{1em}\selectfont}
\renewcommand\arraystretch{1.25}
\setlength{\tabcolsep}{1.2pt}

\begin{table}[t]
    \begin{center}
    \scalebox{0.65}{%
    \hbox{\hspace{0.0 em}
    \begin{tabular}{c|c|c|c}
    \hline
    stage & output & SeqResNet & SeqResNet-B \\ \hline
    conv1 & 32$\times$32 & 3$\times$3, 16 & 3$\times$3, 16 \\ \hline
    \multirow{4}{*}{conv2} & \multirow{4}{*}{32$\times$32} & 3$\times$3, 16$\times r$ $\textcolor{red}{*}$ & 1$\times$1$\to$3$\times$3, 16$\times r$ $\textcolor{red}{*}$ \\ \cline{3-4} 
    \multicolumn{1}{l|}{} & \multicolumn{1}{l|}{} & \blockseqc{16} & \blockseqcb{16} \\
     &  &  &  \\
     &  &  &  \\ \hline
    \multirow{4}{*}{conv3} & \multirow{4}{*}{16$\times$16} & \begin{tabular}[c]{@{}c@{}}3$\times$3, 16$\times r$ \textcolor{blue}{*}\\ Concatenate, 32$\times$r\\ 3$\times$3, 32$\times r$, stride 2, groups = $\frac{32\times r}{k}$\end{tabular} & \begin{tabular}[c]{@{}c@{}}1$\times$1$\to$3$\times$3, 16$\times r$ \textcolor{blue}{*}\\ Concatenate, 32$\times$r\\ 3$\times$3, 32$\times r$, stride 2, groups = $\frac{32\times r}{k}$\end{tabular} \\ \cline{3-4} 
    \multicolumn{1}{l|}{} & \multicolumn{1}{l|}{} & \blockseqc{32} & \blockseqcb{32} \\
     &  &  &  \\
     &  &  &  \\ \hline
    \multirow{5}{*}{conv4} & \multirow{5}{*}{8$\times$8} & \begin{tabular}[c]{@{}c@{}}3$\times$3, 16$\times r$ \textcolor{blue}{*}\\ Concatenate, 48$\times$r\\ 3$\times$3, 48$\times r$, stride 2, groups = $\frac{48\times r}{k}$\end{tabular} & \begin{tabular}[c]{@{}c@{}}1$\times$1$\to$3$\times$3, 16$\times r$ \textcolor{blue}{*}\\ Concatenate, 48$\times$r\\ 3$\times$3, 48$\times r$, stride 2, groups = $\frac{48\times r}{k}$\end{tabular} \\ \cline{3-4} 
    \multicolumn{1}{l|}{} & \multicolumn{1}{l|}{} & \blockseqc{48} & \blockseqcb{48} \\
     &  &  &  \\
    \multicolumn{1}{l|}{} & \multicolumn{1}{l|}{} &  &  \\ \cline{3-4} 
     &  & 1$\times$1, 48$\times r$ & 1$\times$1, 48$\times r$ \\ \hline
     & \multirow{2}{*}{1$\times$1} & global average pool & global average pool \\
     &  & fc, softmax & fc, softmax \\ \hline
    \end{tabular}%
    }
    }
    \end{center}
    \caption{Network template for CIFAR. We use a wide architecture following~\cite{he2016identity}, the widening factor is denoted by $r$. $N$ stands for the number of residual blocks for each stage. SeqConv and WSeqConv layers are marked by \textcolor{red}{*} and \textcolor{blue}{*} respectively.}
    \label{table.1}
    \vspace{-1.5 ex}
\end{table}

\begin{table}[t]
    \begin{center}
    \hbox{\hspace{-1.6 em}
    \scalebox{0.60}{%
    \renewcommand{\blockb}[3]{\multirow{3}{*}{
    \(\left[
    \begin{array}{l}
    \text{1$\times$1, #2}\\
    [-.1em] \text{3$\times$3, #2}\\
    [-.1em] \text{1$\times$1, #1}
    \end{array}\right]\)$\times$#3}
    }

    \renewcommand{\blockx}[3]{\multirow{3}{*}{
    \(\left[
    \begin{array}{l}
    \text{1$\times$1, #2}\\
    [-.1em] \text{3$\times$3, #2, $C$=32}\\
    [-.1em] \text{1$\times$1, #1}\\
    \end{array}\right]\)$\times$#3}
    }

    \renewcommand{\blockdense}[3]{\multirow{3}{*}{
    \(\left[
    \begin{array}{l}
    \text{1$\times$1, #2}\\
    [-.1em] \text{3$\times$3, #1}
    \end{array}\right]\)$\times$#3}
    }

    \renewcommand{\blockseq}[4]{\multirow{3}{*}{
    \(\left[
    \begin{array}{l}
    \text{1$\times$1$\to$3$\times$3, #1, k=#2 \textcolor{blue}{*}}\\
    [-.1em] \text{1$\times$1$\to$3$\times$3, #1, k=#2 \textcolor{blue}{*}}
    \end{array}\right]\)$\times$#3 / $\times$#4}
    }

    \newcommand{\blockseqx}[4]{\multirow{3}{*}{
    \(\left[
    \begin{array}{l}
    \text{1$\times$1$\to$3$\times$3, #1, k=#2, c=#3 \textcolor{blue}{*}}\\
    [-.1em] \text{1$\times$1$\to$3$\times$3, #1, k=#2, c=#3 \textcolor{blue}{*}}
    \end{array}\right]\)$\times$#4}
    }

    \newcommand{\bottleseq}[5]{
        \begin{tabular}[c]{@{}c@{}}1$\times$1$\to$3$\times$3, #1, k=#2 \textcolor{blue}{*}\\ concatenate, #3\\ 3$\times$3, #4, stride 2, #5 groups\end{tabular}
    }

    \newcommand{\bottleseqx}[6]{
        \begin{tabular}[c]{@{}c@{}}1$\times$1$\to$3$\times$3, #1, k=#2, c=#3 \textcolor{blue}{*}\\ concatenate, #4\\ 3$\times$3, #5, stride 2, #6 groups\end{tabular}
    }

    \begin{tabular}{c|c|c|c}
    \hline
    stage & output & \multicolumn{1}{c|}{SeqResNeXt-24} & \multicolumn{1}{c}{SeqResNet B42 / B22} \\ \hline
    conv1 & 112$\times$112 & \multicolumn{2}{c}{\begin{tabular}[c]{@{}c@{}}3$\times$3, 32, stride 2\\ 3$\times$3, 32\end{tabular}} \\ \hline

    \multirow{4}{*}{conv2} & \multirow{4}{*}{56$\times$56} & \multicolumn{1}{c|}{\begin{tabular}[c]{@{}c@{}}1$\times$1$\to$3$\times$3, 256, k=32, c=8 \textcolor{red}{*}\\ 3$\times$3, 256, stride 2, 64 groups\end{tabular}} & \multicolumn{1}{c}{\begin{tabular}[c]{@{}c@{}}1$\times$1$\to$3$\times$3, 128, k=32 \textcolor{red}{*}\\ 3$\times$3, 128, stride 2, 4 groups\end{tabular}} \\ \cline{3-4}
     &  & \multicolumn{1}{c|}{\blockseqx{256}{32}{8}{1}} & \multicolumn{1}{c}{\blockseq{128}{32}{3}{1}} \\
     &  & \multicolumn{1}{c|}{} & \\ 
    \multicolumn{1}{l|}{} & \multicolumn{1}{l|}{} & \multicolumn{1}{c|}{} & \\ \hline
    
    \multirow{4}{*}{conv3} & \multirow{4}{*}{28$\times$28} & \multicolumn{1}{c|}{\bottleseqx{256}{32}{8}{512}{512}{64}} & \multicolumn{1}{c}{\bottleseq{128}{32}{256}{256}{4}} \\ \cline{3-4} 
     &  & \multicolumn{1}{c|}{\blockseqx{512}{64}{8}{1}} & \multicolumn{1}{c}{\blockseq{256}{64}{4}{1}} \\
    \multicolumn{1}{l|}{} & \multicolumn{1}{l|}{} & \multicolumn{1}{c|}{} & \\
     &  & \multicolumn{1}{c|}{} & \\ \hline
    
    \multirow{4}{*}{conv4} & \multirow{4}{*}{14$\times$14} & \multicolumn{1}{c|}{\bottleseqx{512}{64}{8}{1024}{1024}{64}} & \multicolumn{1}{c}{\bottleseq{256}{64}{512}{512}{8}} \\ \cline{3-4} 
     &  & \multicolumn{1}{c|}{\blockseqx{1024}{64}{4}{3}} & \multicolumn{1}{c}{\blockseq{512}{64}{5}{2}} \\
     &  & \multicolumn{1}{c|}{} & \\ 
     &  & \multicolumn{1}{c|}{} & \\ \hline
    
    \multirow{4}{*}{conv5} & \multirow{4}{*}{7$\times$7} & \multicolumn{1}{c|}{\bottleseqx{1024}{64}{4}{2048}{2048}{64}} & \multicolumn{1}{c}{\bottleseq{512}{64}{1024}{1024}{8}} \\ \cline{3-4} 
    \multicolumn{1}{l|}{} & \multicolumn{1}{l|}{} & \multicolumn{1}{c|}{\blockseqx{2048}{128}{4}{1}} & \multicolumn{1}{c}{\blockseq{1024}{128}{3}{1}} \\
     &  & \multicolumn{1}{c|}{} & \\
    \multicolumn{1}{l|}{} & \multicolumn{1}{l|}{} & \multicolumn{1}{c|}{} & \\ \cline{3-4}
     &  & \multicolumn{1}{c|}{1$\times$1, 2048} & \multicolumn{1}{c}{1$\times$1, 1024} \\ \hline
    
    & \multirow{2}{*}{1$\times$1} & \multicolumn{2}{c}{global average pool} \\
    &  & \multicolumn{2}{c}{1000-d fc, softmax} \\ \hline
    
    \multicolumn{2}{c|}{\# params.} & \multicolumn{1}{c|}{\small 26.2$\times$$10^6$} & \multicolumn{1}{c}{\small 25.6$\times$$10^6$ / 11.8$\times$$10^6$} \\ \hline
    \multicolumn{2}{c|}{GFLOPs} & \multicolumn{1}{c|}{\small 4.32} & \multicolumn{1}{c}{\small 5.33 / 2.73} \\ \hline

\end{tabular}%
    }
    }
    \end{center}
    \caption{Network architecture and model complexity of our ImageNet models. SeqResNeXt-24 and SeqResNet-B42 have a model complexity comparable to ResNet-50 while SeqResNet-B22 is about the half model size. SeqConv and WSeqConv layers are marked by \textcolor{red}{*} and \textcolor{blue}{*} respectively.}
    \vspace{-2 ex}
    \label{table.2}
\end{table}

\vspace{-2 ex} \paragraph{Implementation details.}
We build SeqResNet and SeqResNet-B for CIFAR based on~\cite{he2016identity}. The networks are divided into three stages with the feature map sizes of 32$\times$32, 16$\times$16, and 8$\times$8 respectively. We place a regular Conv layer before the first SeqConv layer following~\cite{huang2017densely}, and an equal number of residual blocks for each stage. We insert a down-sampling block between each stage and a 1$\times$1 convolution at the end of the third stage before global average pooling. A classifier consisting of a fully connected layer and a softmax activation is attached after average pooling. The exact specifications of our model template for CIFAR are listed in Table~\ref{table.1}.

For the ImageNet evaluation, we adopt SeqResNet-B and SeqResNeXt with four stages on 224$\times$224 inputs. We use different $k$ for each stage of our ImageNet models due to the increasing model complexity and computational cost. Following~\cite{szegedy2016rethinking, xie2018bag}, We replace the expensive 7$\times$7 convolution and the following max pooling with two 3$\times$3 convolutions and a down-sampling block. We list the detailed configurations of our ImageNet models in Table~\ref{table.2}.

\section{Experiments}
We conduct experiments on three public benchmark datasets: CIFAR-10, CIFAR-100~\cite{krizhevsky2009learning} and ImageNet~\cite{deng2009imagenet}. We compare our models with their original backbones~\cite{he2016deep, xie2017aggregated} and similar state-of-the-art methods~\cite{huang2017densely, zhu2018sparsely}.

\subsection{Datasets}

\paragraph{CIFAR.} 
Both CIFAR-10 (C10) and CIFAR-100~\cite{krizhevsky2009learning} (C100) consist of 50,000 training samples and 10,000 test samples, which are divided into 10 and 100 classes respectively. All samples are color image of 32$\times$32 pixels. We apply widely adopted data augmentation including mirroring and shifting as in~\cite{lin2013network, romero2014fitnets, springenberg2014striving, lee2015deeply, gross2016training} for these two datasets and normalize samples by the channel means and standard deviations. We refer to the augmented datasets as C10+ and C100+. 5,000 training samples are randomly selected for validation as we evaluate our models. We use all training samples for the final run following~\cite{huang2017densely} and report the final test error at the end of training.

\vspace{-2 ex} \paragraph{ImageNet.}
The ImageNet 2012 classification dataset~\cite{deng2009imagenet} contains 1.28 million training images and 50,000 validation images drawn from 1,000 classes. We adopt the standard augmentation scheme following~\cite{gross2016training, xie2017aggregated, huang2017densely} and normalize the dataset by the channel means and standard deviations. We evaluate our models on the single 224$\times$224 center crop following~\cite{he2016identity, xie2017aggregated}.

\subsection{Training}

\begin{figure*}[t]
    \centering
    \includegraphics[width=.39\linewidth]{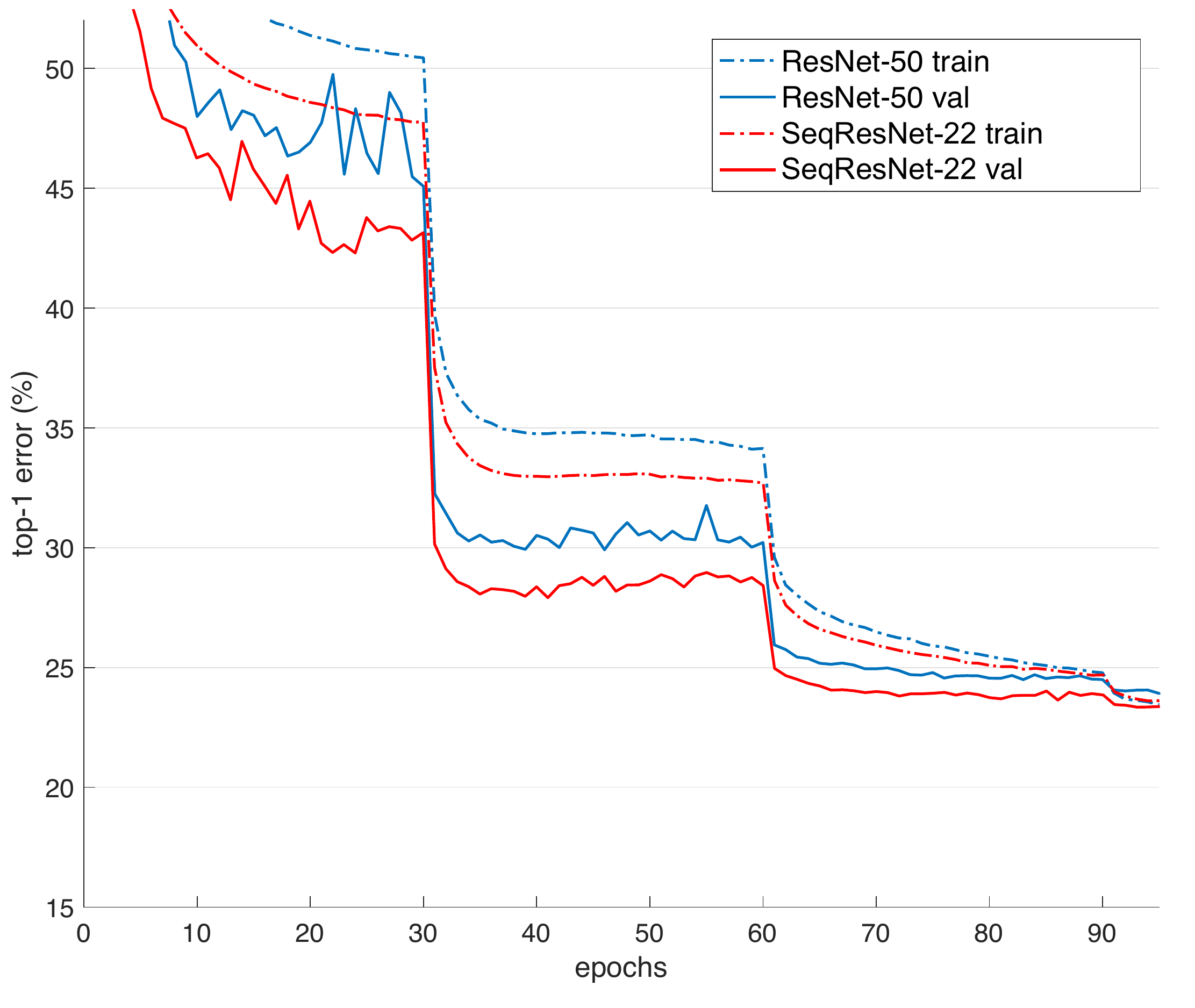}
    \includegraphics[width=.39\linewidth]{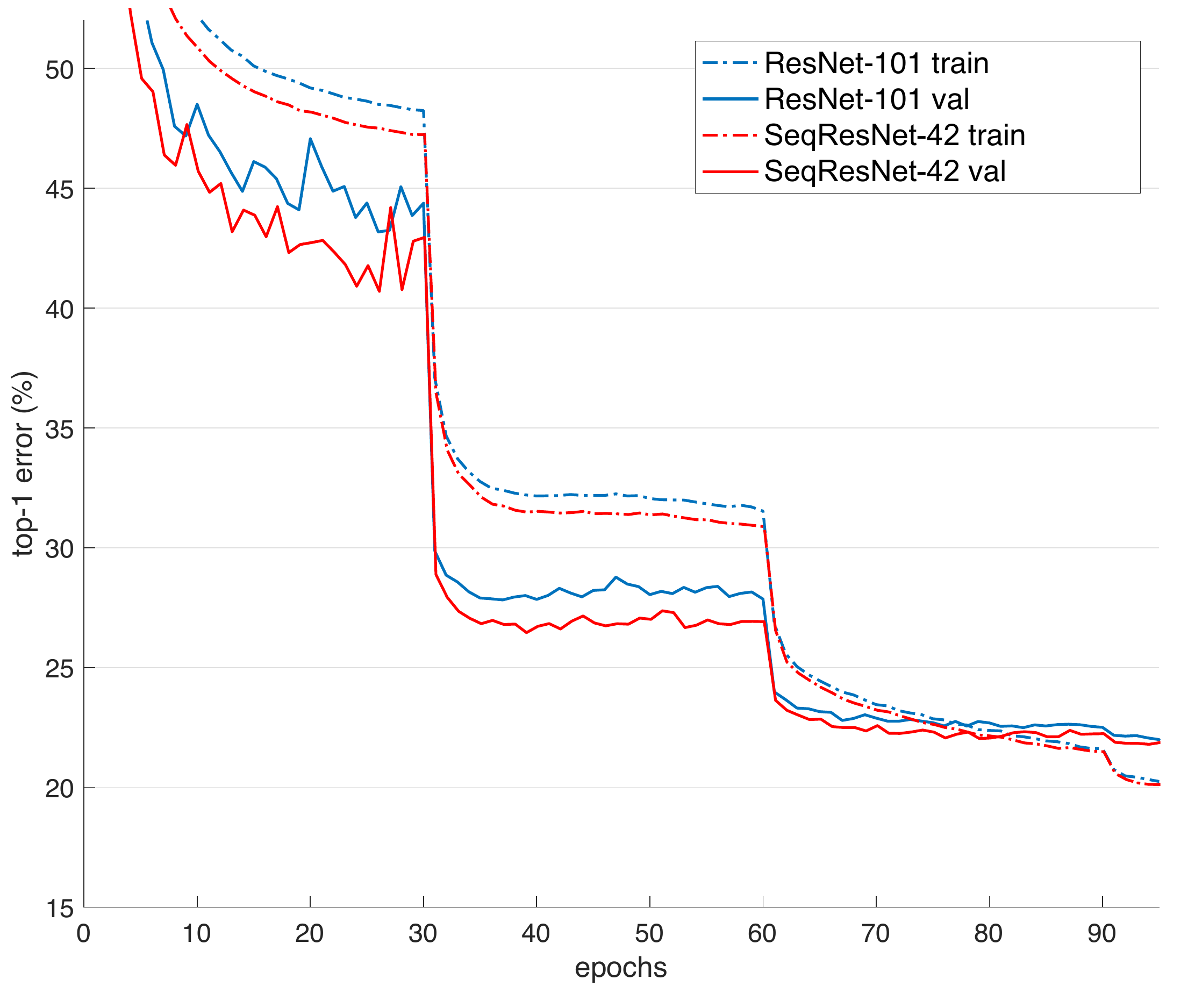}
    \caption{Training curves of SeqResNet-B22/ResNet-50 (left) and SeqResNet-B42/ResNet-101 (right) on ImageNet.}
    \vspace{-1.5 ex}
    \label{fig.4}
\end{figure*}

All models are optimized with stochastic gradient descent (SGD). We apply Nesterov momentum~\cite{sutskever2013importance} of $0.9$ and L2 weight regularization of $10^{-4}$ following~\cite{gross2016training}. We initialize the second WSeqConv layer of each residual block with zeros and all other Conv layers are initialized following~\cite{he2015delving}. The zero initialization disables all residual blocks and imitates a shallow network, which is easier to optimize at the initial stage of the training. Similar initialization procedures for ResNet are also proposed in~\cite{xie2018bag, zhang2019fixup}. We apply the initialization introduced in~\cite{glorot2010understanding} on the fully connected layer of the classifier. The training for all models starts with an initial learning rate of $0.1$.

For the CIFAR datasets, we train our models for $300$ epochs with batch size $64$ and divide the learning rate by $10$ at epoch $150$ and $225$. Due to the limited number of samples presented in these two datasets, we follow~\cite{huang2018condensenet} and apply dropout~\cite{srivastava2014droupout} with a drop rate of $0.1$ before the 1$\times$1 convolution (prior to global average pooling) and every SeqConv layer (except the first one) to reduce overfitting.

The ImageNet models are trained for $100$ epochs with batch size $256$. We report the best validation error of the first $90$ epochs of training and also the best error till all $100$ epochs finish for a fair comparison with~\cite{he2016deep, xie2017aggregated, huang2017densely}. We reduce the learning rate by $10$ times for every $30$ epochs.

\subsection{Results on CIFAR}

\begin{table}[t]
    \begin{center}
    \scalebox{0.65}{%
    \hbox{\hspace{0 em}
    \begin{tabular}{c|c|c|c|c}
    \hline
     & settings &{\# params.}& C10+ & C100+ \\ \hline
    ResNet-110~\cite{he2016deep} & - & 1.7M & 6.61 & - \\ \hline
    ResNet-110 (reported by~\cite{huang2016deep}) & - & 1.7M & 6.41 & 27.22 \\ \hline
    Wide ResNet-16~\cite{zagoruyko2016wide} & 8$\times$ width & 11.0M & 4.81 & 22.07 \\
    Wide ResNet-28 & 10$\times$ width & 36.5M & 4.17 & 20.5 \\ \hline
    ResNet-164 (pre-activation)~\cite{he2016identity} & - & 1.7M & 5.46 & 24.33 \\
    ResNet-1001 (pre-activation) & - & 10.2M & 4.62 & 22.71 \\ \hline
    DenseNet~\cite{huang2017densely} & $L=40$, $k=12$ & 1.1M & 5.24 & 24.42 \\
    SparseNet~\cite{zhu2018sparselya} & $L=40$, $k=12$ & 0.8M & 5.47 & 24.48 \\
    SeqResNet & $k=8$, $r=4$, $N=1$ & 1.2M & \textbf{4.78} & \textbf{22.65} \\ \hline
    DenseNet & $L=100$, $k=12$ & 7.2M & 4.1 & 20.2 \\
    SparseNet & $L=100$, $k=16$, $32$, $64$ & 7.2M & 4.21 & 19.89 \\
    SeqResNet & $k=16$, $r=10$, $N=1$ & 7.6M & \textbf{3.97} & \textbf{19.72} \\ \hline
    DenseNet-BC & $L=100$, $k=12$ & 0.8M & 4.51 & 22.27 \\
    SparseNet-BC & $L=100$, $k=24$ & 1.5M & 4.49 & 22.71 \\
    SeqResNet-B & $k=16$, $r=7$, $N=1$ & 0.8M & \textbf{4.3} & \textbf{20.76} \\ \hline
    DenseNet-BC & $L=100$, $k=16$, $32$, $64$ & 7.9M & 4.02 * & 19.55 * \\
    SparseNet-BC & $L=100$, $k=16$, $32$, $64$ & 4.4M & 4.34 & 19.9 \\
    SeqResNet-B & $k=32$, $r=12$, $N=3$ & 6.0M & \textbf{3.72} & \textbf{18.51} \\ \hline
    \end{tabular}%
    }
    }
    \end{center}
    \caption{Error rates (\%) and model sizes on CIFAR. Results that surpass all competing methods are \textbf{bold}. * indicates results reported by~\cite{zhu2018sparselya}.} 
    \vspace{-1.5 ex}
    \label{table.3}
\end{table}

The experimental results for CIFAR~\cite{krizhevsky2009learning} are presented in Table~\ref{table.3}. SeqResNet outperforms the corresponding ResNet baseline by a large margin. A SeqResNet-B with merely $0.8$M parameters achieves higher accuracies than the 1001-layer ResNet counterpart with more than $10$M parameters, which reduces the model complexity required to obtain an accuracy comparable to that of ResNet by a factor of \textbf{12}. SeqResNet also consistently outperforms the state-of-the-art DenseNet~\cite{huang2017densely} and SparseNet~\cite{zhu2018sparsely} of similar model complexity. A SeqResNet with $1.2$M parameters attains about $0.5\%$ lower error rate on C10+ and $2\%$ lower error rate on C100+ compared to its DenseNet and SparseNet counterparts. SeqResNet shows significantly higher parameter efficiency than wide architectures such as wide ResNet~\cite{zagoruyko2016wide} and ResNeXt\cite{xie2017aggregated} as well. It only takes $6$M parameters for SeqResNet-B to achieve higher accuracies than the $6\times$ larger wide ResNet-28, or attain error rates comparable to the $5.73\times$ larger ResNeXt-29, 8$\times$64d ($34.4$M parameters, $3.65\%$ error rate on C10+ and $17.77\%$ on C100+).

\subsection{Results on ImageNet}
\label{sec:results_on_imagenet}

\begin{table}[t]
    \begin{center}
    \scalebox{0.75}{%
    \begin{tabular}{c|c|c|c|c}
    \hline
     & \# params. & FLOPs & top-1 err & top-5 err \\
    \hline
    ResNet-101~\cite{he2016deep} & 44.5M & 7.34G & 22.44 & 6.21 \\
    DenseNet-264~\cite{huang2017densely} & 33.3M & 5.52G & 22.15 & 6.12 \\
    ResNet-152 & 60.2M & 10.82G & 22.16 & 6.16 \\
    SeqResNet-B42 & 25.6M & 5.33G & \textbf{22.06} & \textbf{5.98} \\
    SeqResNeXt-24 & 26.2M & 4.32G & \textbf{21.92} & \textbf{5.82} \\
    \hline
    ResNet-50 & 25.6M & 3.86G & 24.01 & 7.02 \\ 
    DenseNet-169 & 14.1M & 3.22G & 23.80 & 6.85 \\
    SeqResNet-B22 & 11.8M & 2.73G & \textbf{23.67} & \textbf{6.78} \\ 
    \hline
    ResNet-50 \textcolor{red}{*} & 25.6M & 3.86G & 23.9 & - \\ 
    SeqResNet-B22 \textcolor{red}{*} & 11.8M & 2.73G & \textbf{23.35} & 6.68 \\ 
    \hline
    ResNeXt-50~\cite{xie2017aggregated} \textcolor{red}{*} & 25.0M & 4.00G & 22.2 & - \\
    ResNet-101 \textcolor{red}{*} & 44.5M & 7.34G & 22.0 & - \\
    SeqResNet-B42 \textcolor{red}{*} & 25.6M & 5.33G & \textbf{21.75} & 5.89 \\
    SeqResNeXt-24 \textcolor{red}{*} & 26.2M & 4.32G & \textbf{\textcolor{red}{21.50}} & 5.73 \\ \hline
    \end{tabular}%
    }
    \end{center}
    \caption{Validation error rates on ImageNet. Models marked by \textcolor{red}{*} are trained for 100 epochs.}
    \label{table.4}
    \vspace{-2 ex}
\end{table}

\begin{figure*}[t]
    \begin{center}
        \includegraphics[width=0.95\textwidth]{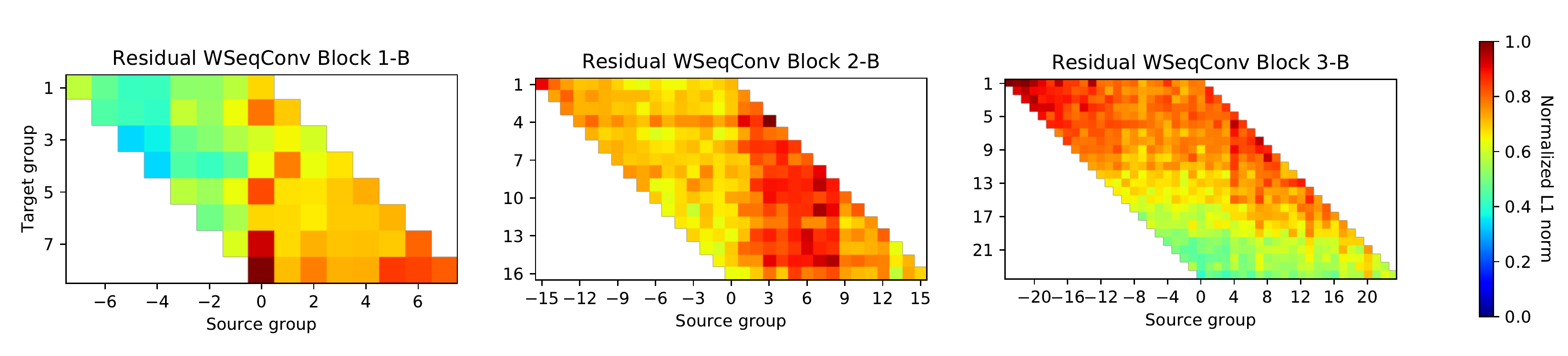}
        \caption{Weight visualization of three WSeqConv layers, we plot the heat map for the second WSeqConv layer of the residual block of each stage from a trained SeqResNet ($k=8$, $r=4$, $N=1$).}
        \label{fig.5}
        \vspace{-3.5 ex}
    \end{center}
\end{figure*}

We evaluate SeqResNet-B and SeqResNeXt on the large-scale ILSVRC 2012 dataset to validate the scalability of our models. Table~\ref{table.4} reports the top-1 and top-5 validation errors of our models on ImageNet. 

SeqResNet-B22 and SeqResNet-B42 not only surpass the performance of their ResNet counterpart of equal model size, but even go much further and outperform ResNets of significantly larger model complexity. Figure~\ref{fig.4} (left) shows that SeqResNet-B22 with less than $12$M parameters exhibits lower training error and validation error than the much larger ResNet-50 with more than 25M parameters. A similar trend between SeqResNet-B42 and ResNet-101 is also plotted in Figure~\ref{fig.4} (right). The lower training error with much smaller model size indicates that SeqResNet has much stronger representational ability than ResNet, in fact, SeqResNet-B42 even outperforms the 2.35$\times$ larger ResNet-152. SeqResNet also shows superior performance compared against the state-of-the-art DenseNet~\cite{huang2017densely} and ResNeXt~\cite{xie2017aggregated}. Both SeqResNet models attain higher accuracy than their DenseNet counterpart with fewer parameters. The performance gap between SeqRetNet-B42 and ResNeXt-50 of similar model complexity, is marked by the $2\times$ complexity ResNet-101 that ResNeXt-50 fails to outperform but surpassed by SeqResNet-B42.

Further performance gain could be observed on SeqResNeXt-24. It has a model complexity similar to SeqResNet-B42 but achieves higher accuracy with lower computational cost (FLOPs). SeqResNeXt-24 also significantly outperforms its ResNeXt-50 counterpart by a showing a top-1 error rate comparable to the $2\times$ complexity ResNeXt-101 ($21.2\%$). To the best of our knowledge, SeqResNeXt-24 has the current best accuracy on ImageNet (with similar augmentation and training/testing procedures) for non-NAS~\cite{zoph2018learning} based models of similar model complexity (about $25$M parameters).

\subsection{Discussion}

\paragraph{Hyperparameter Investigation.}
We empirically evaluate the effect of each hyperparameter as listed in Table~\ref{table.5}. All comparing models have been adjusted to a similar complexity. $S1$ is the standard reference model as we present in Table~\ref{table.3}. $S2$ disables windowed aggregation for all SeqConv layers, it thus has a smaller $r$ (width) than $S1$ since a basic SeqConv layer has a higher complexity than WSeqConv. $S3$ adopts a larger $k$ than $S1$. $S4$ further reduces $r$ (width) and increases $N$ (number of residual blocks) for $S3$. The higher error rate of $S2$ compared to $S1$ verifies the effectiveness of windowed aggregation. $S3$ has marginally more parameters than $S1$ while showing a higher error rate, which indicates that smaller $k$ (more groups, deeper representation) might be beneficial to the representation capability. However, further performance gain on CIFAR with even smaller $k$ is negligible. It is possible that the samples in the CIFAR dataset are too small (32$\times$32) to utilize extremely deep features. Comparison between $S3$ and $S4$ shows that a wide network with fewer residual blocks performs better than a narrow network with more residual blocks.

\vspace{-2 ex} \paragraph{Weight Visualization.}
We conduct the experiment proposed in~\cite{huang2017densely} to visualize the weights of a trained SeqResNet. We pick 3 WSeqConv layers, each from a residual block of SeqResNet trained on C10+. We plot the weights based on the exploded view of SeqConv that the normalized weights corresponding to the connection between two groups are represented by a colored pixel. Results are plotted as heat maps in Figure~\ref{fig.5}. A red pixel indicates heavy use of an aggregated feature group while a blue pixel indicates low usage. Pixels of the white color indicate their corresponding feature group does not participate in the aggregation. We observe from the heat maps that there is hardly any blue pixel and a significant portion of the non-white pixels are red, indicating all parameters are reasonably exploited, whereas DenseNet~\cite{huang2017densely} leaves large blue area on its heat maps. Our observation suggests that our model exhibits low parameter redundancy and fully exploits all aggregated features, which might explain the improved performance that our model attains with compact model size.

\section{Conclusion}

\begin{table}[t]
    \begin{center}
    \scalebox{0.8}{
    \begin{tabular}{c|c@{\hskip 2ex}|c@{\hskip 2ex}|c|c|c|c@{\hskip 1ex}} \hline
       settings & $\hspace{2ex}r$ & $\hspace{2ex}k$  & $N$           & windowed aggregation & \# params. & \hspace{1ex}err \\ \hline
    $S1$ & \hspace{2ex}4 & \hspace{2ex}8  & 1           & \checkmark              & 1.2M   & \hspace{1ex}4.78     \\ \hline
    $S2$ & \hspace{2ex}3 & \hspace{2ex}8  & \{1, 2, 1\} & \ding{55}                    & 1.2M   & \hspace{1ex}4.90     \\ \hline
    $S3$ & \hspace{2ex}4 & \hspace{2ex}16 & 1           & \checkmark                    & 1.3M   & \hspace{1ex}4.99     \\ \hline
    $S4$ & \hspace{2ex}3 & \hspace{2ex}16 & \{1, 2, 2\} & \checkmark                    & 1.3M   & \hspace{1ex}5.15    \\ \hline
    \end{tabular}
    }
    \end{center}
    \caption{Error rates (\%) of models with different hyperparameters on C10+.}
    \label{table.5}
    \vspace{-1.5 ex}
\end{table}

We present a new form of aggregation-based convolutional layer (SeqConv) to enhance the representation capability of a single layer. SeqConv comprises various internal groups that are sequentially aggregated to extract features of various depths, and thus exhibits the benefits of both wide representation and deep representation. We also analyze the relations of SeqConv to DenseNet~\cite{huang2017densely} which bears apparent similarity to our work, but is found to be ultimately different. A windowed aggregation mechanism is proposed as well to address the parameter redundancy of dense aggregation. SeqConv has the same model granularity as a regular convolutional layer and thus could be integrated into a wide variety of backbone networks. We adopt ResNet~\cite{he2016identity} and ResNeXt~\cite{xie2017aggregated} as the backbone networks for our models. Experimental results on image classification indicate that our models with SeqConv significantly outperform their original backbones, and perform favorably against state-of-the-art methods~\cite{xie2017aggregated, huang2017densely, zhu2018sparsely}.

\end{document}